\documentclass{article}
\usepackage{spconf,amsmath,graphicx}

\title{Deep-FSMN for Large Vocabulary Continuous Speech Recognition}
\name{Shiliang Zhang$^{1}$, Ming Lei$^1$,  Zhijie Yan$^1$, Lirong Dai$^2$ }

\address{$^1$Alibaba Inc., P. R. China\\
	$^2$NELSLIP, USTC, P. R. China\\
	\{sly.zsl, lm86501, zhijie.yzj\}@alibaba-inc.com, lrdai@ustc.edu.cn}

\begin{document}
\ninept
\maketitle
\begin{abstract}
In this paper, we present an improved feedforward sequential memory networks (FSMN) architecture, namely Deep-FSMN (DFSMN), by introducing skip connections between memory blocks in adjacent layers. These skip connections enable the information flow across different layers and thus alleviate the gradient vanishing problem when building very deep structure. As a result, DFSMN significantly benefits from these skip connections and deep structure. We have compared the performance of DFSMN to BLSTM both with and without lower frame rate (LFR) on several large speech recognition tasks, including English and Mandarin. Experimental results shown that DFSMN can consistently outperform BLSTM with dramatic gain, especially trained with LFR using CD-Phone as modeling units. In the 2000 hours Fisher (FSH) task, the proposed DFSMN can achieve a word error rate of 9.4\% by purely using the cross-entropy criterion and decoding with a 3-gram language model, which achieves a 1.5\% absolute improvement compared to the BLSTM. In a 20000 hours Mandarin recognition task, the LFR trained DFSMN can achieve more than 20\% relative improvement compared to the LFR trained BLSTM. Moreover, we can easily design the lookahead filter order of the memory blocks in DFSMN to control the latency for real-time applications. 
\end{abstract}
\begin{keywords}
DFSMN, FSMN, LFR, LVCSR, BLSTM
\end{keywords}
\section{Introduction}
\label{sec:intro}
Recently, deep neural networks have become the dominant acoustic models in large vocabulary continuous speech recognition (LVCSR) systems.  Depending on how the networks are connected, there exist various types of deep neural networks, such as feedforward fully-connected neural networks(FNNs) \cite{mohamed2012acoustic,dahl2012context} , convolutional neural networks(CNNs)\cite{abdel2012applying,abdel2014convolutional} and recurrent neural networks (RNNs) \cite{elman1990finding,graves2013speech}. As opposed to FNNs that can only learn to map a fixed-size input to a fixed-size output, RNNs can learn to model sequential data over an extended period of time and store the memory in the network weights, then carry out rather complicated transformations on the sequential data. 
Thereby, researchers have paid more and more attention to RNNs, especially the long short-term memory networks (LSTM) \cite{hochreiter1997long}. It is widely observed that LSTM \cite{sak2014long} and its variations \cite{kalchbrenner2015grid,sainath2016modeling}  can significantly outperform the FNNs on various acoustic modeling tasks. 

While RNNs are theoretically powerful, the learning of RNNs usually relies on the so-called back-propagation through time (BPTT) \cite{werbos1990backpropagation} due to the internal recurrent cycles. The BPTT significantly increases the computational complexity of the learning, and even worse, it may cause many problems in learning, such as gradient vanishing and exploding \cite{bengio1994learning}. As an alternative, some feedforward architecture have been proposed to model the long-term dependency. A straightforward attempt is the so-called unfolded RNN \cite{saon2014unfolded}, where an RNN is unfolded in time for a fixed number of time steps. The unfolded RNN only needs comparable training time as the standard FNNs while achieving better performance than FNNs. Time delay neural network (TDNN)  \cite{peddinti2015time,peddinti2015reverberation,waibel1989phoneme} is another popular feedforward architecture which can efficiently model the long temporal contexts.

Recently,  in \cite{zhang2015feedforward_1,zhang2015feedforward}, we have proposed a simple non-recurrent structure, namely feedforward sequential memory networks (FSMN), which can effectively model long term dependency in sequential data without using any recurrent feedback.  Experimental results on acoustic modeling and language modeling tasks have shown that FSMN can significantly outperform the recurrent neural networks and these models can be learned much more reliably and faster. Furthermore, a variant FSMN architecture namely compact FSMN (cFSMN) is proposed in \cite{zhang2016compact,zhang2017nonrecurrent} to simplify the FSMN architecture and further speed up the learning.  

In this work, based on previous FSMN works and recent works on neural networks with very deep architecture
\cite{srivastava2015highway,he2016deep,zhang2016highway}, we have presented an improved FSMN structure namely Deep-FSMN (DFSMN) by introducing skip connections between memory blocks in adjacent layers. These skip connections enable the information flow across different layers and thus alleviate the gradient vanishing problem when building very deep structure.  Moreover, considering demand of real-world applications, we propose to combine DFSMN with lower frame rate (LFR) \cite{pundak2016lower} technology to speed up decoding  and optimize the DFSMN topology to meet the latency requirement.  In \cite{zhang2016compact,zhang2017nonrecurrent}, the previous FSMN works are evaluated on the popular 300 hours Switchboard (SWB) task. In this work, we try to evaluate the performance of DFSMN on several much larger speech recognition tasks, including English and Mandarin. Firstly, in the 2000 hours English Fisher (FSH) task, the proposed DFSMN with much smaller model size can achieve an absolute 1.5\% word error rate reduction compared to the popular BLSTM. Furthermore, in a 20000 hours Mandarin task, the LFR trained DFSMN have achieved more than 20\% relative improvement compared to the latency controlled BLSTM (LCBLSM) \cite{zhang2016highway,xue2017improving}. More importantly, we can easily design the lookahead filter order of the memory blocks in DFSMN to match the latency demand of real-time applications. In our experiments, a LFR trained DFSMN with 5 frames delay can still outperform the LFR trained LCBLSTM with 40 frames delay.
 

\section{From FSMN to cFSMN}
\label{sec:fsmn_cfsmn}
  \begin{figure}
  	\centering
  	\includegraphics[width=0.9\linewidth, height=4.7cm]{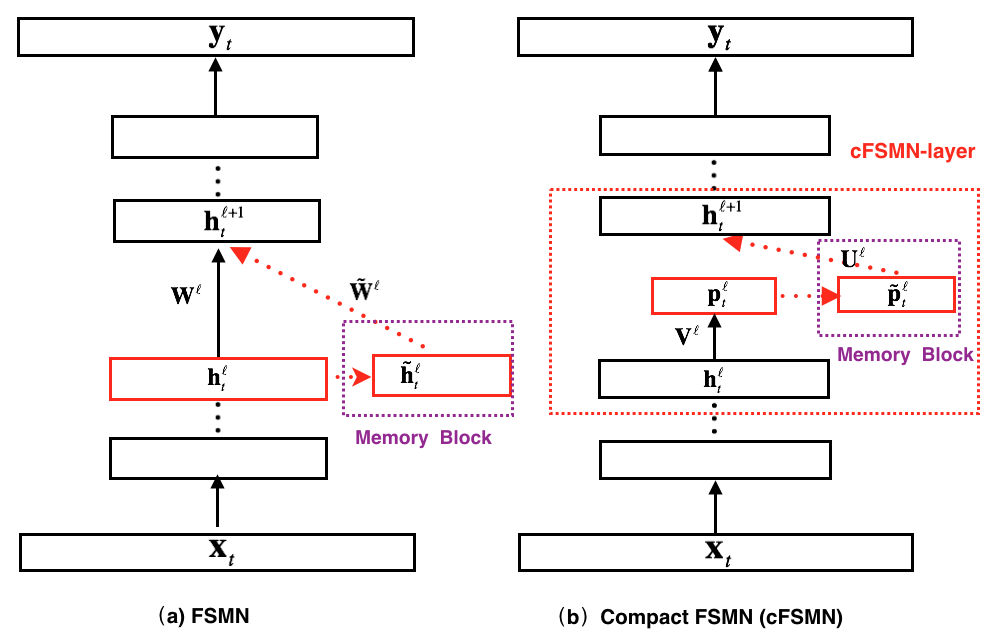}
  	\caption{Illustration of the FSMN and cFSMN. }
  	\label{fig:FSMN_cFSMN}
  \end{figure}

FSMN is proposed in \cite{zhang2015feedforward_1,zhang2015feedforward}, which is inspired by the filter design knowledge in digital signal processing that any infinite impulse response (IIR) filter can be well approximated using a high-order finite impulse response (FIR) filter. Because the recurrent layer in RNNs can be conceptually viewed as a first-order IIR filter, it may be precisely approximated by a high-order FIR filter. Therefore, FSMN extends the standard feedforward fully connected neural networks by augmenting some memory blocks, which adopt a tapped-delay line structure as in FIR filters, into the hidden layers. For instance,  Figure \ref{fig:FSMN_cFSMN}(a) shows a FSMN with one memory block added into its $\ell$-th hidden layer.  The learnable FIR-like memory blocks in FSMNs may be used to encode long context information into a fixed-size representation, which helps the model to capture long-term dependency.  Moreover,  we can add several memory blocks to multiple hidden layers of a deep neural network to capture more context information in various abstraction levels.   In \cite{zhang2015feedforward}, depending on the encoding method to be used, we have proposed two versions of FSMNs, namely scalar FSMNs (sFSMN) and vectorized FSMNs (vFSMN).  For the vFSMN, the formulation of the memory block takes the following form:
 \begin{equation}\label{eq.bi_vector}
 {\bf{\tilde h}}^{\ell}_t = \sum\limits_{i = 0}^{N_1} {\bf a}_i^{\ell}  \odot {\bf h}^{\ell}_{t - i} + 
 \sum\limits_{j = 1}^{N_2} {\bf c}_j^{\ell} \odot {\bf h}^{\ell}_{t + j}  
 \end{equation}
 Where $\odot$ denotes element-wise multiplication of two equally-sized vectors. $N_1$ is called the look-back order,  denoting the number of historical items looking back to the past, and $N_2$ is called the lookahead order, representing the size of the lookahead window into the future. The output from the memory block, ${\bf{\tilde h}}^{\ell}_t$, may be regarded as a fixed-size representation of the long surrounding  context at time instance $t$. As shown in Figure \ref{fig:FSMN_cFSMN} (a), ${\bf{\tilde h}}^{\ell}_t$ can be fed into the next hidden layer in the same way as ${\bf h}^{\ell}_t$.  As a result, we can calculate the activation of the units in the next hidden layer as follows:
 \begin{equation}\label{eq.3}
 {\bf h}_t^{\ell  + 1} = f({\bf W}^\ell  \mathbf{h}_t^\ell  + {{{\mathbf{\tilde W}}}^\ell } {\mathbf{\tilde h}}_t^\ell  + {{\bf b}^\ell })
 \end{equation}

Considering the additional parameters introduced by the memory blocks, a variant FSMN architecture namely compact FSMN (cFSMN) is proposed in \cite{zhang2016compact} to simplify the FSMN architecture and speed up the learning. As shown in Figure \ref{fig:FSMN_cFSMN} (b), it is a cFSMN with a single cFSMN-layer in the $\ell$-th layer. Compared to the standard FSMN, the cFSMN can be viewed as inserting a smaller linear projection layer after the nonlinear hidden layers and adding the memory blocks to the linear projection layers instead of the hidden layers.  The encoding formulation of the memory block in cFSMN takes the following form:
\begin{equation}\label{eq.cfsmn_bi_vector}
{\bf{\tilde p}}^{\ell}_t = {\bf p}^\ell_t+\sum\limits_{i = 0}^{N_1} {\bf a}_i^{\ell}  \odot {\bf p}^{\ell}_{t - i} + 
\sum\limits_{j = 1}^{N_2} {\bf c}_j^{\ell} \odot {\bf p}^{\ell}_{t + j}. 
\end{equation}
Moreover, we can calculate the activation of the units in the next hidden layer as follows:
\begin{equation}\label{eq.3}
{\bf h}_t^{\ell  + 1} = f({\bf U}^\ell {\bf{\tilde p}}^{\ell}_t + {\bf b}^{\ell +1})
\end{equation}

As shown in Figure \ref{fig:FSMN_cFSMN}, both FSMN and cFSMN remain as a pure feedforward structure that can be efficiently learned using the standard back-propagation (BP) with mini-batch based stochastic gradient descent (SGD). 

\section{Deep-FSMN}
\label{sec:DFSMN}
\begin{figure}
	\centering
	\includegraphics[width=0.9\linewidth]{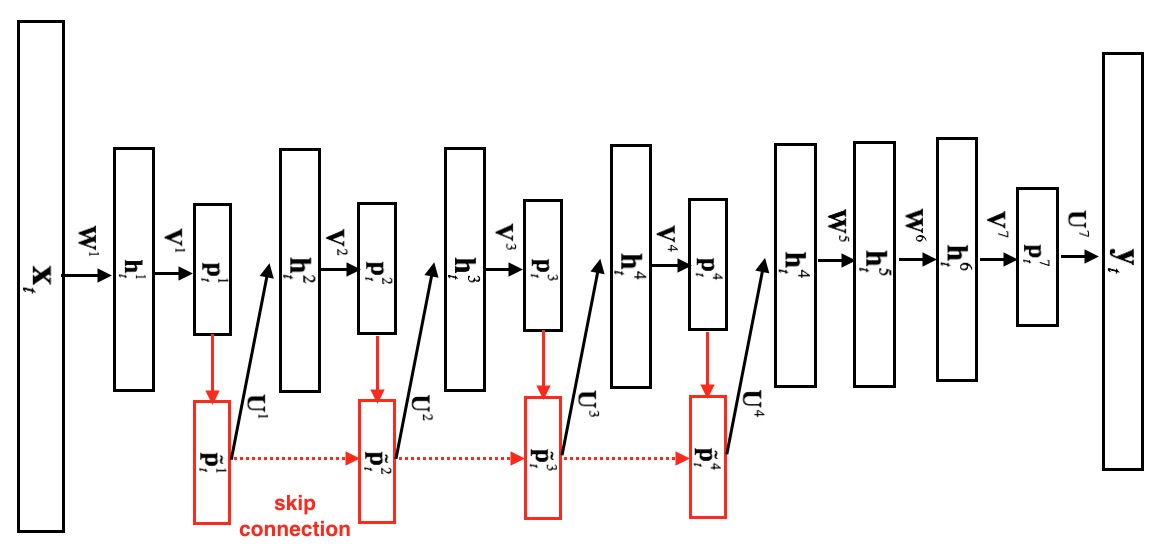}
	\caption{Illustration of Deep-FSMN (DFSMN) with skip connection.}
	\label{fig:DFSMN}
\end{figure}
In previous cFSMN, as introduced in section \ref{sec:fsmn_cfsmn}, the standard hidden layer is decomposed into two layers by using low-rank weight matrix factorization.  Thereby, for a cFSMN with 4 cFSMN-layers and 2 DNN layers, the total number of layers is 12. If we want to train deeper cFSMN by directly adding more cFSMN-layers, it will suffer from the gradient vanishing problem. Inspired by recent works on training very deep neural architectures with skip connection, such as residual \cite{he2016deep}  or highway networks \cite{srivastava2015highway}, we propose an improved FSMN architecture namely Deep-FSMN (DFSMN) in this work .

The architecture of DFSMN is as shown in Figure \ref{fig:DFSMN}. We add some skip connections between the memory blocks of standard cFSMN, where the output of the lower layer memory block can be directed flow to the higher layer memory block. During back-propagation, the gradients of higher layer can also be assigned directly to lower layer that help to overcome the gradient vanishing problem.  The formulation of the memory block in DFSMN takes the following form:
\begin{equation}\label{eq.DFSMN}
{\bf{\tilde p}}^{\ell}_t =\mathcal{H}({\bf{\tilde p}}^{\ell-1}_t)+ {\bf p}^\ell_t+\sum\limits_{i = 0}^{N^\ell_1} {\bf a}_i^{\ell}  \odot {\bf p}^{\ell}_{t - s_1*i} + \sum\limits_{j = 1}^{N^\ell_2} {\bf c}_j^{\ell} \odot {\bf p}^{\ell}_{t + s_2*j}. 
\end{equation}
Here, ${\bf p}^\ell_t={\bf  V}^\ell {\bf h}_t^\ell + {\bf b}^\ell$ denotes the linear output of the $\ell$-th linear projection layer. ${\bf{\tilde p}}^{\ell}_t$ denotes the output of the $\ell$-th memory block. $N^\ell_1$ and $N^\ell_2$ denotes the look-back order and lookahead order of the $\ell$-th memory block, respectively.
$\mathcal{H}(\cdot)$ denotes the skip connection within the memory block, which can be any linear or nonlinear transformation.  For example , if the dimensions of the memory blocks are the same, we can just use the identity mapping as following:
\begin{equation}\label{eq.skip_connection}
{\bf{\tilde p}}^{\ell-1}_t=\mathcal{H}({\bf{\tilde p}}^{\ell-1}_t)
\end{equation}
In this work, we use the identity mapping for all experiments.

For a speech signal, the information of adjacent frames have strong redundancy due to the overlap. Similar to the dilated convolutional layer in wavenet \cite{oord2016wavenet}, we add the stride factors to the memory block in order to remove this redundancy. As in Eq.(\ref{eq.DFSMN}), $s_1$ and $s_2$ are the stride for look-back and lookahead filters respectively. For DFSMN, the total latency ($\tau$)  is relevant to the lookahead filters order (${N}_2^{\ell}$) and the stride ($s_2$) in each memory block, which can be calculated as Eq.(\ref{eq.delay}).
\begin{equation}\label{eq.delay}
    \tau =\sum\limits_{\ell = 1}^{L} {N}_2^{\ell}\cdot s_2
\end{equation}
When speech recognition system is applied to some real-time applications, it is essential to control the latency. For DFSMN, we can easily design the lookahead filters order and the stride to meet the demand of latency. 
In our experiments, we will evaluate the performance of DFSMN with different latency.


\section{Experiments}
\label{sec:exp}
In this section, we evaluate the performance of the proposed DFSMN with lower frame rate (LFR) on several large vocabulary speech recognition tasks including English and Mandarin.

\subsection{English Recognition Task}
 \begin{table}
 	\centering
 	\caption{Performance (WER\%) of various acoustic models trained with CE criterion on the 2000 hours Fisher task. (166$^*$ is an estimate model size based on the configuration in \cite{xiong2016achieving})}
 	\begin{tabular}[t]{c|c|c}
 		\hline
 		Model         & Size (MB)   & WER ( \%) \\\hline
 		DNN     &     159                  &  14.3  \\
 		BLSTM        &     180                 &   \textbf{10.9} \\
 		BLSTM(6)\cite{xiong2016achieving}& 166$^*$ &  \textbf{10.3}\\
 		cFSMN        &     104                  & 10.8  \\     
 		DFSMN(12) &     152                &  \textbf{9.4}    \\\hline
 	\end{tabular}
 	\label{tab:FSH_Dif_AM}
 \end{table}
For the English recognition task, we use the standard Fisher (FSH) task \cite{cieri2004fisher}. The training set consists of  about 2000 hours data from both the Switchboard (SWB) and Fisher (FSH). Evaluation is performed in term of word error rate (WER) on the Switchboard part of the standard NIST 2000 Hub5 evaluation set (containing 1831 utterances), denoted as Hub5e00.  The input speech sampled at 8kHz is analyzed using a 25ms Hamming window with a fixed 10ms frame shift.  We computed the 72-dimensional filter-bank (FBK) features, which includes 24 log energy coefficients distributed on a mel scale, along with their first and second temporal derivatives.  All experiments in this task, we use a tri-gram language model (LM) that is trained on 3 million words from the SWB training transcripts and 11 million words of the Fisher English Part 1 transcripts. 

For the hybrid DNN-HMM baseline system, we follow the same training procedure as described in \cite{zhang2015rectified} to train the conventional context dependent DNN-HMM using the tied-state alignment obtained from a MLE trained GMM-HMM baseline system. The DNN contains 6 hidden layers with 2,048 rectified linear units (ReLU) per layer. The inputs are the stacked FBK feature with a context window size of 15 (7+1+7). For the hybrid  BLSTM-HMM baseline system, we have trained a deep BLSTM by following the same configurations in \cite{sak2014long}. The BLSTM consists of three BLSTM layers (1024 memory cells for each direction) and each BLSTM layer is followed by a low-rank linear recurrent projection layer of 512 units. As to the cFSMN based system, we have trained a cFSMN with architecture being $ 3*72 \textnormal{-} 4 \times[2048 \textnormal{-}512(20,20)] \textnormal{-} 3 \times  2048 \textnormal{-} 512 \textnormal{-} 9004 $.  The inputs are the 72-dimensional FBK features with context window being 3 (1+1+1). The cFSMN consists of 4 cFSMN-layers followed by 3 ReLU DNN hidden layers and a linear projection layer.

 All models are trained in a distributed manner using BMUF\cite{chen2016scalable} optimization on 8 GPUs and frame-level cross entropy criterion. The initial learning rate is 0.00001, and the momentum is kept as 0.9. For DNN and cFSMN, the mini-batch is set to be 4096.
 BLSTM model is trained using the standard full-sequence BPTT with a mini-batch of 16 sequences.  The performances of the baseline models are as shown in Table \ref{tab:FSH_Dif_AM}. 

We have trained DFSMN with various architectures, which can be denoted as  $3*72\textnormal{-}N_f\times[2048\textnormal{-}512(N_1;N_2;s_1;s_2)]\textnormal{-}N_d\times2048\textnormal{-}512\textnormal{-}9004$. Here $N_f$ and $N_d$ are the number of cFSMN-layer and ReLU DNN layer respectively.  In these experiments, $N_1=20, N_2=20, N_d=3$  is kept fixed.  In the first experiment, we have investigated the influence of the number of cFSMN-layers and the size of the stride on the
final speech recognition performance.  We have trained cFSMN with six, eight, ten and twelve cFSMN-layers. Detailed architectures
and experimental results are listed in Table \ref{tab:Dif_DFSMN}. Here, \emph{DFSMN(6)} denotes DFSMN with $N_c$ being 6.  Results of \emph{exp1} and \emph{exp2} indicate the advantage of using stride for the memory block. From \emph{exp2} to \emph{exp5}, we can achieve consistent performance improvement by using deeper architecture.

In Table \ref{tab:FSH_Dif_AM}, we have summarized experimental results of various systems on the 2000 hours task. Our implementation of the BLSTM achieve a WER of 10.9\% which is about 23.8\% relative improvement compared to the baseline DNN system. For comparison, a well-trained BLSTM with six hidden layers (512 cells per direction) in \cite{xiong2016achieving}  achieves a WER of 10.3\% by decoding with a 4-gram language model. As to the proposed DFSMN, we can achieve a WER of 9.4\%  by purely using the cross-entropy criterion without any feature space or speaker space adaptation technologies, which is a very competitive performance in this task.  Compared to our baseline BLSTM system, the proposed DFSMN can achieve an absolute 1.5\% WER reduction with a smaller model size. 

\begin{table}
	\centering
	\caption{Performance (WER\%) of DFSMN based acoustic model with various architectures.}
	\begin{tabular}[t]{c|c|c|c|c}
		\hline
ID&	Model   & stride & Size(MB) & WER (\%) \\\hline
exp1&	DFSMN(6) &  1   & 104      & 10.7 \\
exp2&   DFSMN(6) & 2    & 104 & 10.3\\
exp3&    DFSMN(8) & 2   & 120  & 9.6\\
exp4&    DFSMN(10) & 2  & 136 & 9.5\\
exp5&    DFSMN(12) & 2  & 152  & 9.4\\\hline
	\end{tabular}
	\label{tab:Dif_DFSMN}
\end{table}

\subsection{Mandarin Recognition Task}
For the Mandarin recognition task, we have evaluated the performance of the proposed DFSMN on two tasks, namely 5000-hour-task and 20000-hour-task, which consist of 5000 hours and 20000 hours training data respectively. The 5000-hour-task is a subset of the full 20000-hour-task. The training data are collected from many domains, such as sport, tourism, game, literature et al. A test set contains about 30 hours data is used to evaluate the performance of all models. Evaluation is performed in term of character error rate (CER in \%).  The sample rate of the data is 16kHz. Acoustic feature used for all experiments are 80-dimensional log-mel filterbank (FBK) energies computed on 25ms window with 10ms shift. 

\subsubsection{5000-hour-task}
\begin{table}[t]
	\centering
	\caption{Comparison of various acoustic models on 5000-hour-task. }
	\begin{footnotesize}
		\begin{tabular}[t]{c|c|c|c|c}
			\hline
			Model               & Target          & Size (MB)  & CER \% & Gain \\\hline
			LCBLSTM           & CD-State            &  196         & 18.78         & -    \\
			cFSMN(6)           &          &  102         & 17.72         & +5.32\% \\\hline 
			\bf{LFR-LCBLSTM}&    CD-Phone   &  \bf{220} &  \bf{18.92}         & - \\\hline
			LFR-cFSMN(6)       &      & 108          &  16.85         & +11.00\% \\
			LFR-cFSMN(8)       &   CD-Phone   & 124          &  15.80         & +16.50\% \\
			LFR-cFSMN(10)       &     & 140          &  15.91         & +15.86\% \\\hline
			LFR-DFSMN(8)       &       & 124          &  15.45        & +18.34\% \\
			\bf{LFR-DFSMN(10)}& CD-Phone   & \bf{140}    &  \bf{15.00} & \bf{+20.72\%} \\\hline             
		\end{tabular}
	\end{footnotesize}
	\label{tab:5000hour_task}
\end{table}
\begin{table}[t]
	\centering
	\caption{Comparison of LFR trained LCBLSTM and DFSMN on training time and real-time factor (RTF).}
	\begin{tabular}[t]{c|c|c}
		\hline
		Model                   & Training time (hr/epoch) & RTF \\\hline
		LFR-LCBLSTM       &   21.62                 &   0.4289 \\\hline
		LFR-DFSMN(8)      &    6.85                  &   0.1486  \\\hline             
	\end{tabular}
	\label{tab:5000hour_task_efficient}
\end{table}

In this experiments , we evaluate the performance of DFSMN with CD-state and CD-Phone as modeling units. For comparison, we have also trained the latency controlled BLSTM (LCBLSM)\cite{zhang2016highway,xue2017improving} as our baseline systems. For the CD-Phone models, we use the lower frame rate (LFR) \cite{pundak2016lower} technology with frame rate being 30ms. 

For the conventional hybrid models with CD-state, an existing cross-entropy trained hybrid CD-DNN-HMM system is used to realign and generate the new 10ms frame-level targets.  The HMM consists of 14359 CD-states. For the baseline hybrid CD-LCBLSTM-HMM system, we follow the well-tuned configuration as in \cite{xue2017improving} to train a LCBLSTM with $N_c$ and $N_r$ being 80 and 40 respectively. The baseline LCBLSTM consists of 3 BLSTM layers (500 memory cells for each direction), 2 ReLU DNN layers (2048 hidden nodes for each layer) and a softmax output layer.  As to the cFSMN based model, we have trained a cFSMN with architecture being $ 3*80 \textnormal{-} 6 \times[2048 \textnormal{-}512(20,20)] \textnormal{-} 2 \times 2048 \textnormal{-} 512 \textnormal{-} 14359 $.   The inputs are the 80-dimensional FBK features with context window being 1 and 3 for LCBLSTM and cFSMN respectively. 

As to the LFR trained hybrid models with CD-Phone, we firstly map the 14359 CD-states to 9841 CD-Phones and then subsample by averaging 3 one-hot target labels (LFR is 30ms), producing soft LFR targets. For the baseline LFR trained LCBLSTM system (LFR-LCBLSTM), we use a similar model architecture to the baseline system while with $N_c$ and $N_r$ being 27 and 13 respectively.
For LFR trained cFSMN models (denoted as LFR-cFSMN), we have trained cFSMNs with six, eight and ten cFSMN-layers, denoted as \emph{LFR-cFSMN(6)}, \emph{LFR-cFSMN(8)} and \emph{LFR-cFSMN(10)} respectively. The inputs are the 80-dimensional FBK features with context window being 17 and 11 for LCBLSTM and cFSMN respectively.  For the LFR trained DFSMN model (LFR-DFSMN), the model topology is denoted as $11*80\textnormal{-}N_f\times[2048\textnormal{-}512(N_1;N_2;s_1;s_2)] \textnormal{-}N_d\times2048\textnormal{-}512\textnormal{-}9841$.  In these experiments, we set $N_1=10,N_2=5,s_1=2,s_2=2,N_d=2$, and then try to evaluate the performance of  LFR-DFSMN with $N_c$ being 8 and 10, denoted as \emph{LFR-DFSMN(8)} and \emph{LFR-DFSMN(10)} respectively.

All models are trained in a distributed manner using BMUF\cite{chen2016scalable} optimization on 8 GPUs and frame-level cross entropy criterion. We have listed all experimental results in Table \ref{tab:5000hour_task} for comparison.  Compared to the baseline CD-state trained LCBLSTM, the cFSMN can achieve 5.32\% relative improvement.  The LFR-cFSMN with CD-Phone can achieve about 0.9\% absolute CER reduction compared to the cFSMN with CD-state, which shows the benefit from the modeling units. Both cFSMN and DFSMN can benefit from the deep architecture, and DFSMN can outperform the cFSMN with the similar model topology. However, too deep cFSMN (such as 10 layers) will suffer from the performance degradation while the DFSMN can achieve consistent improvement. Moreover, the proposed LFR-DFSMN can significantly outperform the LFR-LCBLSTM with about 20\% relative CER reduction. We may train much deeper LCBLSTM to achieve better performance, such as the Highway-LSTM in \cite{pundak2017highway}. However, the improvement is limited if without increasing the total parameters, which only about 2\% relatively improvement. In Table \ref{tab:5000hour_task_efficient}, we have compared the training time and decoding real-time factor of LFR trained DFSMN and LCBLSTM. Results shown that DFSMN can achieve about 3 times speedup in training and decoding real-time factor (RTF).

\subsubsection{20000-hour-task}
\begin{table}[t]
	\centering
	\caption{Comparison of various acoustic models on 20000-hour-task.(``1 and 0" denotes the lookahead filter order of the odd layer and even layer is 1 and 0 respectively).}
		\begin{footnotesize}
	\begin{tabular}[t]{c|c|c|c|c}
		\hline
         Model              & $N_2$  & Delay Frame & CER\%  & Gain \\\hline
         LFR-LCBLSTM   &    -        &  40              & 16.05   & -\\\hline
                                &    2       &   20              & 12.67    &+21.06\% \\
         LFR-DFSMN(10)&    1        &   10              & 12.94 & +19.38\%\\
                                &    1 and 0 &  5               & 13.38  &+16.64\%\\\hline 
               
	\end{tabular}
	\end{footnotesize}
	\label{tab:2000hour_task}
\end{table}
In this task, we try to compare the performance of LFR-LCBLSTM and LFR-DFSMN on a very large corpus that consists of 20000 hours training data. For LFR-LCBLSTM, we use the same configurations to the 5000-hour-task. 
For LFR-DFSMN, we have trained DFSMN with model topology being $11*80-10\times[2048-512(5;N_2;2;1)]\textnormal{-}2\times2048\textnormal{-}512\textnormal{-}9841$, denoted as \emph{LFR-DFSMN(10)}.
We fix the number of FSMN layers($N_f$), DNN layers ($N_d$), look-back filter order ($N_1$) and try to investigate the influence of lookahead filter order ($N_2$) to the performance.   All models are trained in a distributed manner using BMUF\cite{chen2016scalable} optimization on 16 GPUs and frame-level cross entropy criterion.  

For the baseline LFR-LCBLSTM with $N_c=27$ and $N_r=13$, the number of delay frame for time instance is 40. For LFR-DFSMN, we can control the number of delay frame by setting the  lookahead filter order. Experimental results in Table \ref{tab:2000hour_task} have shown that when reduce the number of delay frame from 20 to 5, we only loss about 5\% performance. As a result, the latency is about 150ms ($30ms*5$) which is suitable for real-time applications. Finally, the proposed DFSMN with 20 frames latency can achieve more than 20\% relative improvement compared to the LCBLSTM with 40 frames latency.

\section{Conclusions}
We presented an improved FSMN structure namely Deep-FSMN (DFSMN), and applied it to many large speech recognition tasks. The DFSMN can significantly benefit from the skip connections and the deeper architecture. Experimental results shown that the DFSMN consistently outperform the BLSTM with dramatic gain, especially when combined with the lower frame rate. In the 2000 hours Fisher English task, our proposed DFSMN can achieve a WER of 9.4\% by purely using cross-entropy criterion without any adaptation technology.  In a 20000 hours Mandarin recognition task, the LFR trained DFSMN can achieve more than 20\% relative improvement compared to the LFR trained LCBLSTM while with smaller model size and lower latency. Experiments results suggest that the DFSMN may be a strong alternative to BLSTM for acoustic modeling.

\vfill\pagebreak

\bibliographystyle{IEEEbib}
\bibliography{refs}

\end{document}